\documentclass[10pt,twocolumn,letterpaper]{article}

\usepackage{cvpr}      %

\usepackage{iftex}
\ifXeTeX\usepackage[T1]{fontenc}\fi

\usepackage{cuted} %

\usepackage{currfile} %

\usepackage{caption} %

\usepackage{multirow}
\usepackage{pifont}
\newcommand{\cmark}{\textcolor{green!55!black}{\ding{51}}}
\newcommand{\xmark}{\textcolor{red!80!black}{\ding{55}}}
\usepackage{tikz}
\usetikzlibrary{positioning,arrows.meta,fit,backgrounds,calc,shapes.geometric,shapes.misc}

\definecolor{cvprblue}{rgb}{0.21,0.49,0.74}
\usepackage[pagebackref,breaklinks,colorlinks,allcolors=cvprblue]{hyperref}

\def\confName{CVPR}
\def\confYear{2027}

\newcommand{\x}{\mathbf{x}}
\newcommand{\R}{\mathbf{R}}
\newcommand{\tvec}{\mathbf{t}}
\newcommand{\ba}{\boldsymbol{\theta}}
\newcommand{\bb}{\boldsymbol{\beta}}
\newcommand{\uu}{\mathbf{u}}
\newcommand{\X}{\mathbf{X}}

\title{Part-wise Certified State Estimation of Articulated Objects from Passive Stereo}

\author{Orion Debonnet$^{1}$, Kei Higuchi$^{2}$, Alex Sykes$^{1}$, Ayush Deshmukh$^{1}$, Victor Leve$^{1}$, Joao Moura$^{1}$, Sethu Vijayakumar$^{1}$\\
$^{1}$University of Edinburgh \quad $^{2}$Kawada Robotics}

\begin{document}
\twocolumn[{%
\renewcommand\twocolumn[1][]{#1}%
\maketitle
\begin{center}
  \centering
  \captionsetup{type=figure}
  \includegraphics[width=\textwidth]{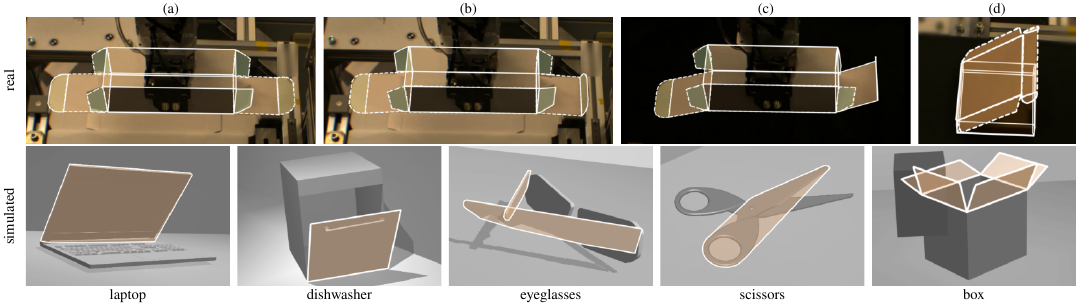}
  \vspace{-1.5em}
  \captionof{figure}{\textbf{Our estimates on hinged objects made of thin parts.} Top: three phases (a, b, c) during folding of a real carton on the robot jig, and a smaller test carton (d). Bottom:
  five simulated PartNet-Mobility objects, shown in grey levels. Moving parts are filled, with a solid outline where our certificates
  accept the part and a dashed one where they do not.}
  \label{fig:teaser}
\end{center}%
}]

\begin{abstract}
Robots in manufacturing often handle thin, hinged objects known from CAD models
that change shape during the task, like folding cartons. Monitoring such tasks
requires estimating the object's articulation state, its pose and the shape of any semi-deformable part. Learning-based estimators
generalise across objects but lose precision while handling thin parts. Model-based fitting is precise
but often needs accurate depth, a close initial estimate and a clean object mask. We present part-wise certification of an
articulated object's state, estimated from passive stereo. Our estimator rivals the precision of the best model-based fitting method without using depth, a close initial estimate or a mask. We detect the corners that the CAD model defines in both stereo images and fit the articulated model to them. Geometric certificates, which we extend from rigid to articulated objects, then tell whether to trust each part, because the fit returns a state even for hidden parts. Certified estimates on
unlabelled video become labels that improve sim-to-real transfer. On five of seven simulated objects,
we locate the parts we certify three to five times more precisely than the strongest learned baseline on the same readings,
given depth and true masks. On a real folding carton, our median corner error on the end panels is 2.1\,\% of the carton's size,
at least 40\,\% lower than every learned baseline and also lower than a depth-based fit initialised near the true state.
\end{abstract}

\section{Introduction}
\label{sec:intro}

As robots take over more assembly and packaging, they increasingly perform tasks that bring an object from an initial
articulation state to a goal state through manipulation, as when they fold a carton or close a lid. Many objects are made of thin, nearly flat parts joined by hinges,
like a carton's flaps or a laptop's screen. Supervising the task requires the object's \emph{articulation
state}, its pose, hinge angles and the shape of any semi-deformable part. A failure leaves the object in a state the task
should not reach, with a part folded the wrong way, out of order or not far enough. The state reveals such failures and
properties that vary between items, such as the stiffness of a crease. Manufactured
objects come with a computer-aided design (CAD) model, so their articulated shape is known and only their state
needs measuring. A failure often shows in a single part, so every part needs a precise estimate and a test of whether
that estimate can be trusted, since no ground truth exists while the robot works.

Meeting both requirements from images is difficult, for four reasons. (a) The parts are thin sheets of board,
metal or plastic. Depth sensors are least reliable on them, at the edges that decide the state, and without depth,
their lack of texture leaves little to match between two images. (b) The state has many coupled hinges,
and an error at one hinge shifts every part beyond it. (c) The manipulator and the object itself occlude parts, and near
symmetries make the visible parts ambiguous. (d) Few applications provide labelled real images of their object, and networks trained only
on renders fail on real images, often confidently. The first two difficulties limit precision, and the last
two limit trust.

Existing methods fall into two families with opposite strengths. Learning-based estimators predict the state with
networks trained on many objects, either for a category of articulated
objects~\cite{li2020ancsh,geng2023gapartnet,capnet2025,dicart2026} or for any rigid object given its CAD
model~\cite{labbe2022megapose,wen2024foundationpose}. They generalise across objects and tolerate partial views. In
our experiments, however, they lose precision on thin parts, because they replace the exact articulated model with a
learned shape or a rigid one. Model-based fitting instead adjusts a known kinematic model until it matches the measured
depth~\cite{michel2015kinematic,desingh2019factored,schmidt2014dart}. With accurate depth, a clean object mask and a close initial estimate,
usually the previous frame's state, it can be very precise. However, depth errors can be as large as the gaps between thin parts, similar objects nearby can spoil the mask, and after an occlusion
the last estimate may be far from the true state. We therefore seek the precision of model-based fitting from a calibrated stereo pair of colour cameras, with
no close initial estimate and no object mask.

\begin{figure*}[t]
  \centering
  \resizebox{0.95\textwidth}{!}{\input{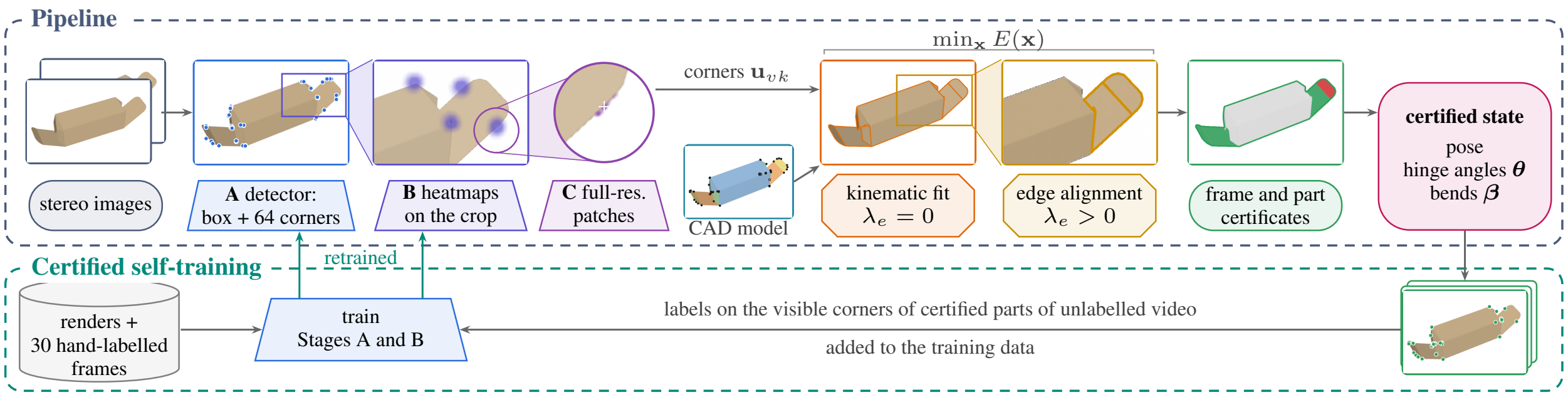}}
  \caption{\textbf{Overview.} Networks locate the corners coarse to fine in both images. A kinematic fit adjusts the
  carton's pose, hinge angles and panel bends to them, then to the panels' edges. Geometric certificates decide
  which estimates become labels for the next round of training. Shown on a render, with the networks' real outputs for Stages A to C. Green parts pass the certificate, the red one fails.}
  \label{fig:pipeline}
\end{figure*}

To this end, we use the corners of the object's articulated CAD model as correspondences. Networks detect these
corners coarse to fine in both images of a passive stereo pair, and a kinematic fit adjusts the model to them (\cref{fig:pipeline}). Each
corner is a known point of the model, so the two views need no texture to be matched, as Yu~\etal~\cite{boss2025}
showed for rigid objects. Fitting the model to both views at once triangulates the corners, so it needs neither a depth sensor nor a disparity map. The fit starts from several
generic states and stays within the hinge limits, so it needs no close initial estimate. Through the model, the
visible corners of a part also constrain its hidden ones. The approach therefore needs parts with distinct corners,
which thin hinged parts usually have.

A fully hidden part, however, is not constrained at all, yet the fit still returns a state for it. Geometric
certificates~\cite{talak2023c3po,boss2025} guard against such issues. A certificate tests an estimate against the
measurements alone, for instance whether each detected corner lies where the fitted model projects it, so it runs on
every frame at inference time. Existing certificates have been applied only to rigid
objects, and they accept or reject an estimate as a whole. We certify each part of an articulated object separately.
The certificates do not refine the estimate. They decide which parts to trust and to track, and which frames to learn from. Certified estimates on unlabelled video become training labels, which adapt the
networks to real images with only a few hand-labelled frames. Unlike~\cite{boss2025}, we certify each part, train only on visible corners and hold hidden parts at their last certified angle.

We study one instance in depth, a folding carton with eleven hinges (\cref{fig:teaser}, top), on real sequences with multi-camera ground
truth. Five simulated PartNet-Mobility objects~\cite{xiang2020sapien}, a laptop, a dishwasher, a pair of
eyeglasses, a pair of scissors and a box (\cref{fig:teaser}, bottom), test the transfer to other objects given only their CAD models, and a second, smaller carton the transfer to another real object. On all of them but the box, our method estimates the hinge angles of the main moving parts with a lower median error than learned estimators given depth and
true masks, which otherwise do better only on the carton's small tabs. In summary, we contribute (i) a method that recovers the articulation state of an object made of thin, textureless
parts from passive stereo, by fitting its articulated CAD model to detected corners, then to its edges
(\cref{sec:corners,sec:solver,sec:edges}), (ii) to our knowledge, the first part-by-part certification of an articulated
object, and self-training on its certified estimates, with labels restricted to the visible corners
(\cref{sec:certificates,sec:selftrain}), and (iii) an evaluation on real and simulated data
against category-level, render-and-compare and model-based methods given depth, including depth computed from our
stereo cameras, and ground-truth masks (\cref{sec:experiments}).

\section{Related Work}
\label{sec:related}

\paragraph{Articulated object pose.}
Category-level methods estimate the part poses and joint states of unseen instances from
depth~\cite{li2020ancsh,weng2021captra,xue2021omad,geng2023gapartnet,capnet2025,dicart2026} or from a single RGB image or
video~\cite{zhang2025r2art,zhang2026trackarticulate}. They learn the shape of a category instead of using an exact model, and they are
designed for objects with a few joints. With a known kinematic model, the state can instead be fitted to depth, for
articulated, possibly deforming objects~\cite{michel2015kinematic,desingh2019factored,schmidt2014dart,walsman2017deformable} and for
hands~\cite{tagliasacchi2015articulatedicp}. Robots are tracked from colour images alone by
render-and-compare~\cite{labbe2021robopose,liu2024drrobot}, which relies on the distinct shape and appearance of their
links. Like these model-based methods, we
fit a known kinematic model, but to corners detected in a stereo pair.

\paragraph{Pose from a CAD model.}
Render-and-compare methods~\cite{labbe2022megapose,wen2024foundationpose} estimate the pose of a novel rigid object
from its CAD model. Applied to an articulated object part by part, these
methods lose the coupling between parts, and a single thin panel is nearly symmetric. Keypoint
methods~\cite{peng2019pvnet} instead detect points defined on the model and solve for the pose geometrically.
KeyPose~\cite{liu2020keypose} detects them in stereo and halves the error of a single view, even against a baseline
given true depth. Yu~\etal~\cite{boss2025} recover the pose and size of rigid boxes from corners detected in stereo.
Model-based tracking with moving edges~\cite{bouthemy1989moving,drummond2002realtime} locates straight edges more
precisely than points. We combine stereo corners and edges on an articulated model.

\paragraph{Certificates.}
Certifiable algorithms prove that a geometric estimate is globally optimal. They exist for relative
pose~\cite{briales2018certifiably}, rotation averaging~\cite{eriksson2018rotation}, point cloud registration with
outliers~\cite{yang2021teaser}, shape from 2D landmarks~\cite{yang2020perfectshape}, and the category-level pose and
shape of an object~\cite{shi2021pace}, which CAST~\cite{shaikewitz2024cast} extends to tracking a rigid object over
time. An optimal estimate can still be wrong, however. Pose estimators have therefore long verified their hypotheses
against the measurements, by global hypothesis verification in point clouds~\cite{aldoma2012global} or by learned
scores that rank pose hypotheses~\cite{okorn2021zephyr,wen2024foundationpose}. Conformal prediction bounds the pose
error of keypoint-based estimators~\cite{yang2023conformal}, and introspective perception learns to predict when a
vision system fails~\cite{daftry2016introspective}. Certificates of observable correctness check whether an estimate
agrees with the measurements. C-3PO~\cite{talak2023c3po} introduced them for keypoint-based pose. Later work uses them
to choose among an ensemble of estimators~\cite{shi2023ensemble}, to filter pseudo-labels for test-time
adaptation~\cite{shi2025crisp}, and to verify box estimates from stereo~\cite{boss2025}. All of these accept or reject
an estimate as a whole, and those made for objects assume them rigid, possibly of variable shape. We certify an
articulated object part by part, and each part's certificate decides whether it may guide the next frame.

\paragraph{Self-training and adaptation.}
Self-supervised 6D pose methods~\cite{wang2020self6d,chen2023texpose,tan2023smocnet} fine-tune a network trained on
renders with unlabelled real images, and camera relocalisers adapt on the fly to a new scene~\cite{cavallari2017onthefly}.
Certified self-training~\cite{talak2023c3po,shi2023ensemble,boss2025} keeps only the estimates that pass their
certificates as labels. All of these address rigid objects or scenes. We bring certified self-training to an
articulated object, whose labels must follow what the images show of each part (\cref{sec:selftrain}).

\paragraph{Folding and manipulation.}
Robotic carton folding plans the motion from the known
blank~\cite{liu2002carton,yao2008dexterous,yao2011automatic,preziosa2025boxfolding} and executes it without measuring
the carton. Closest to our goal, Yang~\etal~\cite{yang2011boxpackaging} label the planes of a depth point cloud as the
sides and flaps of an open box, coarsely and once, before the motion. Box perception for logistics handles closed,
rigid boxes~\cite{boss2025,box6d2025} or the straps around them~\cite{liang2025straps}. Folded paper and cloth are perceived without an exact model, from corners,
planes or meshes fitted to images or
depth~\cite{choi2024paperfolding,he2020flatdeformation,namiki2021origami,huang2022medor,wang2024trtm,xue2023garmenttracking}.
Digital twins and simulators model the physics of deformable objects~\cite{jiang2025phystwin} and of folding
cardboard~\cite{zhang2026boxtwin,shrivastava2026cardboard}, and they need the object's state over time, which our method measures.
Under occlusion by the manipulator, state estimation can add touch~\cite{delaguila2024visuotactile} or segment and
track the occluding bodies~\cite{long2021rigidfusion}. We instead certify which parts the cameras constrain. To our
knowledge, no prior work tracks the full fold state of a carton through its folding, or certifies it.

\section{Method}
\label{sec:method}

\subsection{State definition}
\label{sec:problem}
\label{sec:model}

We describe an object by its CAD model $\mathcal{M}$, a kinematic tree of rigid parts joined by revolute joints, which we
call hinges, each with limits $[l_h,u_h]$. We cut semi-deformable parts into rigid strips. The model carries $K$ corners, and
corner $k$ sits at a fixed position $\mathbf{p}_k$ in the frame of its part or strip $b(k)$. The articulation state $\x$
places each corner in the reference camera:
\begin{equation}
  \x = (\R, \tvec, \ba, \bb), \quad \X_k(\x) = \R\,\mathbf{T}_{b(k)}(\ba,\bb)\,\mathbf{p}_k + \tvec,
  \label{eq:fk}
\end{equation}
where $(\R,\tvec)$ is the pose of the root part, $\ba$ the hinge angles, $\bb$ the bending coefficients, and
$\mathbf{T}_{b(k)}$ the rigid transform from $b(k)$ to the root part along the kinematic chain. In a semi-deformable part
$p$, the angle between strips $j$ and $j{+}1$ follows a low-order Legendre profile,
$\phi_j = \bar\phi_j + \sum_{m<M} \beta_{pm} P_m(s_j)$, where $s_j\in[-1,1]$ is the joint's position along the part and
$\bar\phi_j$ its rest angle. Given $V\geq2$ calibrated cameras, we estimate $\x$ and decide which of its parts can be
trusted. The CAD model defines what the networks measure, which states the fit may reach and what the certificates check.

The carton (\cref{fig:carton}) has eleven hinges and $K{=}64$ corners. Its two side panels are semi-deformable, with two
bending coefficients each ($M{=}2$), and every other panel is rigid. Turned end for end, the carton nearly maps onto itself, an ambiguity we resolve along tracks (\cref{sec:selftrain}).
\subsection{Coarse-to-fine corners}
\label{sec:corners}

Corners are where a thin, textureless object still has contrast, but rounded creases and occlusion make them hard to
place. We locate them in three stages, run independently on each image. \emph{Stage A}, a Keypoint
R-CNN~\cite{he2017maskrcnn}, finds the object and gives a first estimate of every corner at a fraction of the camera's
resolution. \emph{Stage B} crops the object's bounding box and predicts one heatmap per corner with a Simple Baselines
head~\cite{xiao2018simple}. A heatmap gives a sub-pixel detection $\uu_{vk}$ of corner $k$ in view $v$ and, from its peak, a
confidence $c_{vk}$, and the corner counts as visible, $k\in\mathcal{V}_v$, if $c_{vk}>c_{\min}$. One heatmap pixel spans about
nine camera pixels, so \emph{Stage C}
revisits each corner in a $96{\times}96$ window at full resolution. It predicts the heatmaps of all corners in the window, reads
only the queried one and discards a move of more than 8 pixels, so it refines a corner but never swaps it for another. We train
it only on exact corners, from renders and hand-labelled frames, because the self-training labels (\cref{sec:selftrain}) are
the pipeline's own detections and would teach it the offsets it exists to remove.

\subsection{Kinematic fit}
\label{sec:solver}

The detections are independent and noisy, and the model ties them together. One energy defines the estimate,
\begin{equation}
  \begin{split}
  E(\x) = {}& \sum_{v}\sum_{k\in\mathcal{V}_v} \rho_\sigma\!\big(\lVert \pi_v(\X_k(\x)) - \uu_{vk}\rVert\big) \\
           & + \lambda_\theta \sum_{h\in\mathcal{H}} \big(\theta_h - \theta_h^{\star}\big)^2
             + \lambda_\beta \lVert \bb \rVert^2 \\
           & + \lambda_e \sum_{v,i} \rho_{\sigma_e}\!\big(\mathbf{n}_{vi}^{\top}\big(\pi_v(\mathbf{P}_i(\x)) - \mathbf{m}_{vi}\big)\big),
  \end{split}
  \label{eq:solver}
\end{equation}
subject to $\theta_h\in[l_h,u_h]$ and $|\beta_{pm}|\leq\beta_{\max}$, where $\pi_v$ projects into view $v$. The first term is the corner term, with the Geman-McClure loss~\cite{geman1987statistical}
$\rho_\sigma(r)=\sigma^2r^2/(\sigma^2+r^2)$, and the last is the edge term of \cref{sec:edges}. We minimise $E$ twice, first with $\lambda_e=0$. On video, the second term pulls each hinge of $\mathcal{H}$, those certified since the track began, towards its last certified
angle $\theta_h^\star$ (\cref{sec:certificates}). The third keeps semi-deformable parts at their rest shape unless the images
say otherwise. A stiff penalty closes kinematic loops, such as the glued seam of the carton's tube.

\begin{figure}[t]
  \centering
  \includegraphics[width=\linewidth]{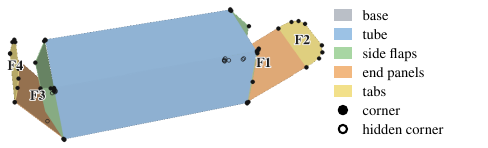}\\[2pt]
  \includegraphics[width=0.9\linewidth]{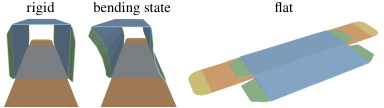}
  \caption{\textbf{Carton model.} Top: the tube open and the end panels and side flaps part-folded, with its eleven hinges
  and 64 corners, hollow where hidden. F1 and F3 are the end panels, F2 and F4 their tabs. Bottom: the tube seen end on,
  rigid and in one bending state of its side panels, and the flat carton, whose layers lie within 6\,mm, less than the
  depth camera's 10\,mm offset (\cref{sec:exp_real}).}
  \label{fig:carton}
\end{figure}

A hinge angle trades off against the root pose and its neighbours, so we solve in two phases. Adam with graduated
non-convexity~\cite{yang2020gnc}, which shrinks $\sigma$ as it iterates, runs from several starts, among them the previous
frame, the flat object and its 180$^\circ$ turns. Levenberg-Marquardt then polishes the start with the lowest energy, where Adam
alone leaves hinge errors of several degrees. Both phases enforce the hinge and bending limits.

\subsection{Edge alignment}
\label{sec:edges}

A rounded, often hidden corner can only be placed so well, whereas a straight edge can be located across its direction far
more precisely. We take samples $\mathbf{P}_i(\x)$ along each straight edge of the fitted model, drop those the object hides,
and project them into both views. Along the normal $\mathbf{n}_{vi}$ of each sample, we search for the strongest image edge of
the same orientation, as in moving-edges tracking~\cite{bouthemy1989moving,drummond2002realtime}, at position
$\mathbf{m}_{vi}$. We then refit the
state with $\lambda_e>0$ and $\lambda_\theta=\lambda_\beta=0$ in \cref{eq:solver}. The corner term keeps the fit from sliding along a line or jumping to a
neighbouring edge. Only parts with visible corners contribute edges, and hinges
the corners leave unconstrained stay fixed, so a hidden part is never dragged onto a background line. We compute the frame certificate
(\cref{sec:certificates}) before this step, so edges cannot accept a fit the corners rejected.

\subsection{Certificates}
\label{sec:certificates}

A network can be confidently wrong, so we test each estimate before trusting it, following the correct-and-certify
approach~\cite{talak2023c3po,boss2025}. A certificate is a binary test of the estimate against the images, calibration and model, so it needs no ground truth. Under occlusion, a frame can pass such tests
while a hidden part sits anywhere within its hinge limits, so we certify each frame and then each of its parts. On our real test
session, 29 end panels are read more than 10\,mm off. The frame certificate accepts 12 of them, and the part certificate none
(see supplementary material).

A visible corner $k$ is certified in view $v$ if it passes two tests from~\cite{boss2025}. The \emph{residual} test,
$\lVert\pi_v(\X_k(\hat\x))-\uu_{vk}\rVert<\epsilon_r$ for the fitted state $\hat\x$, rejects detections the model
cannot explain. The \emph{epipolar} test, $d_F(\uu_{lk},\uu_{rk})<\epsilon_e$ for a corner visible in both views,
bounds the distance of each detection from the epipolar line of the other.

A frame is certified if it passes two tests. \emph{Support}, at least $n_s$ certified corners in each view, ensures that the
measurements determine the state, like the non-degeneracy certificate of C-3PO~\cite{talak2023c3po}. \emph{Scale}, which we add,
compares the certified corners triangulated from the two views with the fitted ones. The median ratio $\kappa$ of their
distances from their centroids must satisfy $|\kappa-1|<\epsilon_s$, which catches disparity errors that the epipolar test
cannot see.

A part $p$ of a certified frame is certified if at least $\min(3,K_p)$ of its $K_p$ corners are certified in either
view and the median residual $r_p$ of its corners visible in both views stays below $\epsilon_p$. We set $\epsilon_r{=}4$,
$\epsilon_e{=}2$ and $\epsilon_p{=}3$ pixels and $\epsilon_s{=}0.05$ on earlier recordings and renders, before we labelled the test
session, and use them for every object. The support $n_s$ is 12 on the carton and 4 on the other objects, which have 7 to 24 corners. A certified part
sets $\theta_h^\star$ of its hinge to the fitted angle, and an uncertified one leaves it unchanged. A fully hidden
part, which no corner constrains, is therefore held at its last certified state by \cref{eq:solver}, and its guessed
angle never steers later fits.

\subsection{Certified self-training}
\label{sec:selftrain}

Certified estimates on unlabelled real video adapt the detector from renders to real images. As in~\cite{boss2025}, they
become labels for Stages A and B, alongside the renders and only 30 hand-labelled frames of the carton. An articulated, partly hidden object
needs two changes to this scheme.

First, a label may teach only what the images show. We label a corner hidden when the fitted model hides it behind one of the object's own parts,
because otherwise the detector learns hidden corners only from renders and reports them confidently on real
images. We label every other certified corner visible at its detection and ignore the rest.

Second, we resolve near symmetries along tracks. The object turns little between frames, whereas an end swap is a
180$^\circ$ jump. We anchor each track on certified frames whose orientation matches the known layout of the workspace,
then give every certified frame the way round closer to its neighbour's. The same step orients the estimates we report.
Tracks cannot resolve an exact symmetry. When a half turn of the CAD maps the object onto itself and validation renders show
the detector confusing the two ends, the detector instead learns corner names relative to the camera, and we read the state up to
that turn. Among our objects, only the box needs this.

\section{Experiments}
\label{sec:experiments}

All experiments use one passive stereo pair, real or simulated, with
1936$\times$1096-pixel images and a 14\,cm baseline.

\subsection{Baselines and metrics}
\label{sec:setup}

\begin{table}[t]
  \centering
  \footnotesize
  \setlength{\tabcolsep}{3pt}
  \begin{tabular}{lcccc@{\hspace{8pt}}l}
    \cmidrule[\heavyrulewidth](r{2pt}){1-5}\cmidrule[\heavyrulewidth](l{2pt}){6-6}
    \multirow{2}{*}{method} & \multicolumn{4}{c}{works without} & \\
    \cmidrule(lr){2-5}
                   & depth & mask & start & CAD & trained on \\
    \cmidrule[\lightrulewidth](r{2pt}){1-5}\cmidrule[\lightrulewidth](l{2pt}){6-6}
    CAP-Net        & \xmark & \xmark & \cmark & \cmark & own renders \\
    DICArt         & \xmark & \xmark & \cmark & \cmark & our renders \\
    FoundationPose & \xmark & \xmark & \cmark & \xmark & generic renders \\
    DART           & \xmark & \xmark & \xmark & \xmark & none \\
    ours           & \cmark & \cmark & \cmark & \xmark & our renders$^*$ \\
    \cmidrule[\heavyrulewidth](r{2pt}){1-5}\cmidrule[\heavyrulewidth](l{2pt}){6-6}
  \end{tabular}
  \caption{\textbf{What each method works without} at inference time (\cmark), and what it is trained on. $^*$On the real carton, also 30
  hand-labelled frames, certified labels from unlabelled video, and the workspace layout to tell the carton's ends apart.}
  \label{tab:inputs}
\end{table}

We compare with both families of \cref{sec:intro} (\cref{tab:inputs}). CAP-Net~\cite{capnet2025} and
DICArt~\cite{dicart2026} estimate the parts of objects of a known category, and FoundationPose~\cite{wen2024foundationpose}
registers a rigid CAD model to each part by render and compare. The model-based tracker
DART~\cite{schmidt2014dart} fits the articulated CAD to the depth inside the object's mask. DART needs a start close to the
true state, which only the truth can provide here, so we start it from the truth perturbed by 5\,mm, 3$^\circ$ and
5$^\circ$ per hinge, or from our answer.

Every baseline receives the input it was designed for, plus true masks, and DART its start near the truth, neither available in deployment, so their results show each method at its best. We aim for comparable precision without these inputs, not for better precision than depth methods in optimal conditions. In simulation, the baselines' depth is exact. CAP-Net solves a different problem, unseen instances of a known category without a CAD, and its training code is not released, so
we run its released weights and refine its estimate against the part's CAD (CAP-Net + CAD). We retrain DICArt on our renders with
the authors' code, which does not reach its published errors here (see supp.), and without the ground-truth refinement of its released evaluation. Where a reading is ambiguous, as for a thin
plate turned by 180$^\circ$, we score the most favourable one.

\subsection{Simulation: the carton and six other objects}
\label{sec:exp_sim}
\label{sec:exp_objects}
\label{sec:exp_carton_baselines}

We test our method on simulated versions of the two cartons of \cref{sec:exp_real} and on five PartNet-Mobility~\cite{xiang2020sapien} objects, chosen for their different difficulties, the
first four from the test split of ArtImage~\cite{xue2021omad}. The laptop's screen is a thin plate, the
dishwasher's door a large plate that its body often hides, the eyeglasses' legs small parts on two hinges, the
scissors' blade a narrow part, and the box's four flaps look alike under a half turn. The carton has 300 test frames and every other object 500, with randomized views, hinge angles and lighting. Other objects hide at most 40\,\% of the object and of each part, and each part shows at least a tenth of its area. Our only object-specific input is the CAD. We derive its corners automatically, as the nearest vertex to each corner of each part's bounding box or, on the small carton, each board's sharp corners, and run only Stages A and B (\cref{fig:pipeline}), without edge alignment, trained on renders alone and with the box's corners named as in \cref{sec:selftrain}.

\Cref{tab:sim} scores each reading, one moving part in one frame, by the distance between the centres of its estimated and true bounding boxes and by its hinge angle.
Neither error depends on how a method defines the part's frame. The supplementary material adds the angle between the estimated and true planes of the part. On the
carton, the moving parts are the end panels F1 and F3, which close the tube, and the smaller tabs F2 and F4 that hang from them
(\cref{fig:carton}). The column ``ours, cert. parts'' keeps only the readings whose part passes its certificate
(\cref{sec:certificates}), 45 to 86\,\% per row. Every method answers over 90\,\% of the readings, except CAP-Net + CAD (65 to 100\,\%) and DICArt on the carton (31\,\%).

\begin{table}[t]
  \centering
  \footnotesize
  \setlength{\tabcolsep}{1pt}
  \resizebox{\linewidth}{!}{%
  \begin{tabular}{lcccccc@{\hspace{6pt}}c}
    \cmidrule[\heavyrulewidth](r{2pt}){1-7}\cmidrule[\heavyrulewidth](l{2pt}){8-8}
         & CAP-Net &        & Foundation- & ours,  & ours, cert. & FP, same & DART, \\
         & + CAD   & DICArt & Pose        & all    & parts       & readings & from truth \\
    \cmidrule[\lightrulewidth](r{2pt}){1-7}\cmidrule[\lightrulewidth](l{2pt}){8-8}
    \multicolumn{8}{l}{\emph{centre of the moving part (mm)}} \\
    end panels & -- & 73/110 & \underline{0.99}/17 & 1.4/22 & 1.1/\textbf{3.2} & \textbf{0.87}/\underline{8.3} & 0.69/1.4 \\
    tabs       & -- & -- & \underline{0.45}/15 & 1.5/19 & 1.1/\underline{3.7} & \textbf{0.36}/\textbf{1.6} & 0.64/1.5 \\
    laptop     & 19/64 & 50/95 & 5.3/18 & \underline{1.4}/\underline{15} & \textbf{1.2}/\textbf{3.9} & 5.0/18 & 3.4/8.3 \\
    dishwasher & 56/217 & 559/888 & 8.7/47 & \underline{5.1}/559 & \textbf{2.8}/\textbf{9.8} & 8.4/\underline{41} & 1.3/3.0 \\
    eyeglasses & -- & 41/71 & 2.1/20 & \underline{0.54}/\underline{4.8} & \textbf{0.44}/\textbf{1.1} & 2.3/23 & 0.57/3.6 \\
    scissors   & -- & 32/50 & 1.3/3.4 & \underline{0.46}/\underline{1.5} & \textbf{0.43}/\textbf{1.1} & 1.3/2.8 & 0.47/3.3 \\
    box        & 6.1/105 & -- & 1.5/60 & \underline{1.4}/\underline{37} & \textbf{1.2}/\textbf{2.9} & 1.4/47 & 0.74/1.0 \\
    small carton & 46/70 & -- & 3.1/20 & \underline{0.73}/\underline{2.6} & \textbf{0.62}/\textbf{1.8} & 2.8/15 & 0.87/2.1 \\
    \cmidrule[\lightrulewidth](r{2pt}){1-7}\cmidrule[\lightrulewidth](l{2pt}){8-8}
    \multicolumn{8}{l}{\emph{hinge angle ($^\circ$)}} \\
    end panels & -- & 21/61 & 1.5$^\dagger$/8.0 & \underline{0.91}/21 & \textbf{0.70}/\textbf{2.3} & 1.2$^\dagger$/\underline{4.7} & 0.43/1.3 \\
    tabs       & -- & -- & \underline{1.3}/26 & 2.3/46 & 1.9/\textbf{6.1} & \textbf{1.2}/\underline{6.4} & 0.83/3.3 \\
    laptop     & -- & 21/56 & 3.7/55 & \underline{0.25}/\underline{10} & \textbf{0.21}/\textbf{0.99} & 4.0/56 & 1.0/2.5 \\
    dishwasher & -- & 22/52 & 0.79/13 & \underline{0.27}/34 & \textbf{0.12}/\textbf{0.58} & 0.73/\underline{3.4} & 0.12/0.28 \\
    eyeglasses & -- & 20/48 & 0.84/6.8 & \underline{0.18}/\underline{2.8} & \textbf{0.12}/\textbf{0.47} & 0.87/7.7 & 0.19/1.4 \\
    scissors   & -- & -- & 1.6/12 & \underline{0.15}/\underline{0.68} & \textbf{0.14}/\textbf{0.45} & 1.2/12 & 0.08/0.34 \\
    box        & -- & -- & 0.59/16 & 0.67/32 & \underline{0.56}/\textbf{2.3} & \textbf{0.51}/\underline{5.3} & 0.14/0.46 \\
    small carton & -- & -- & 5.3/50 & \underline{0.84}/\underline{4.7} & \textbf{0.61}/\textbf{2.6} & 4.6/44 & 1.9/8.0 \\
    \cmidrule[\heavyrulewidth](r{2pt}){1-7}\cmidrule[\heavyrulewidth](l{2pt}){8-8}
  \end{tabular}}
  \caption{\textbf{Simulation}, median/90th percentile, with symmetric parts pooled and the box read up to its half turn. $^\dagger$Base read through the top panel, the tube's true shape given. FP, same readings: FoundationPose on the readings we certify. Readings answered, plane errors and CAP-Net without the CAD: supplementary. Bold: best per row, underlined: second best, DART set apart.}
  \label{tab:sim}
\end{table}

With exact depth and a start near the truth, DART leads on the carton, the dishwasher and the box, matches our medians
on the eyeglasses and scissors, and trails ours on the laptop and the small carton. On the carton, from our solver's generic starts, with the same depth, its corners are a median
13\,mm off, against 1.3\,mm for ours on the same frames. Started from our answer, it lowers our errors by about a third (see supp.), so accurate depth
complements our method.

FoundationPose places the end panels at least as well as we do (1.0 against 1.4\,mm) and the tabs better (0.45 against
1.5\,mm), since exact depth and a true mask nearly determine a flat part. Neither error sees a half turn of a flat part about one of its axes, which leaves its centre and plane unchanged. FoundationPose's corners are 53\,mm off as output and 1.4\,mm after the half turn that best matches the truth. DICArt cannot represent eleven hinges, so it receives one end panel and the panel
it hangs from, and it misses their position by 7\,cm. We leave out CAP-Net, which
has never seen a carton (see supp.).

On the laptop, eyeglasses, scissors and small carton, we lead on position and hinge, and we locate the part three to four times more
precisely than FoundationPose, the closest baseline. On the dishwasher, we lead only on the medians, as a quarter of our fits turn its near-cubic body, nearly all rejected by the part certificate (see supp.). On the readings we certify, FoundationPose's error on these five objects is three to five times ours. The box is the exception. Its medians nearly tie with FoundationPose's, which reads its hinges slightly better. On the
readings we certify, our tail is far shorter, 2.9 against 47\,mm. The part certificate rejects 72 to 100\,\% of readings that are more than 10\,mm off.

\subsection{Fold state of real cartons}
\label{sec:exp_real}

A robot or a person folds the carton ($150\times54\times54$\,mm when folded, 168\,mm across) about 0.8\,m from a stereo pair
placed like a robot's head cameras. Two more static cameras are used only to label the ground truth. As networks trained on renders alone fail on real images (\cref{sec:ablations}), training
adds 30 hand-labelled real frames, 59 views in all, and the certified labels of the same three sessions' unlabelled video. We recorded the test session on another
day and never trained on it. Its 47 hand-labelled frames come from three recordings, and the carton moves in 29 of them. The
recordings belong to an industrial partner, and we cannot release them.

To compare each method with the ground truth, we label each test frame's corners by hand in all cameras and triangulate them. We score millimetres only on the 327 of 997 corners seen by three or more cameras, where a pixel of labelling error means about 0.4\,mm of position error. The CAD fitted to these corners gives the end panels' signed angle to the base, our measure of the fold.
The baselines receive depth from semi-global matching~\cite{hirschmuller2008stereo} on our pair (3\,mm off, a third of the carton covered), a learned stereo network~\cite{wen2025foundationstereo} on the same
pair (2\,mm off, all covered) and an Intel RealSense D435 beside it (76\,\% covered, 10\,mm off).

\begin{table}[t]
  \centering
  \footnotesize
  \setlength{\tabcolsep}{1.5pt}
  \begin{tabular}{lllcccc}
    \toprule
    & & & & end panels & tabs & fold \\
    method & depth & truth & frames & (mm) & (mm) & ($^\circ$) \\
    \midrule
    CAP-Net          & stereo  & M & 8  & 89 / 112 & -- & -- \\
    DICArt           & D435    & M & 18 & 57 / 110 & -- & 64 / 129 \\
    FoundationPose   & learned & M & 47 & 6.3 / 76 & \textbf{2.7} / 144 & 7.5$^\dagger$ / 83 \\
    FoundationPose   & learned & M$^\ddagger$ & 43 & 27 / 146 & 32 / 122 & 21$^\dagger$ / 88 \\
    DART             & learned & M, S & 47 & 6.7 / \textbf{19} & 6.7 / \textbf{19} & \underline{4.7} / \textbf{15} \\
    DART             & learned & S$^\ddagger$ & 44 & 131 / 159 & 139 / 158 & 22 / \underline{64} \\
    \midrule
    ours, single frame & --    & -- & 45 & \underline{4.1} / 99 & 6.4 / 90 & 4.8 / 167 \\
    ours               & --    & -- & 45 & \textbf{3.5} / \underline{59} & \underline{5.4} / \underline{46} & \textbf{3.9} / 82 \\
    \quad certified parts & -- & -- & 33 & 3.7 / 6.6 & 5.2 / 7.1 & 3.7 / 7.1 \\
    \midrule
    \multicolumn{7}{l}{\emph{small carton: corners of the moving parts and their hinge angles}} \\
    CAP-Net          & learned & M & 12 & 38 / 119 & -- & -- \\
    FoundationPose   & learned & M & 12 & 8.6 / 85 & -- & 11 / 68 \\
    DART             & learned & M, S & 12 & \underline{4.7} / \underline{9.5} & -- & \textbf{1.8} / \textbf{12} \\
    ours               & --    & -- & 12 & \textbf{3.5} / \textbf{9.2} & -- & \underline{4.0} / \underline{21} \\
    \quad certified parts & -- & -- & 11 & 1.5 / 4.8 & -- & 1.1 / 7.6 \\
    \bottomrule
  \end{tabular}
  \caption{\textbf{Real cartons}, median / 90th percentile of the error on corners seen by three or more cameras and on the fold, over the frames each method answers. Small carton: its lid and flaps, not the semi-oval flaps, which no corner pins. Truth given at inference: M, true masks, S, a start near the truth on every frame. $^\ddagger$Once per recording, not scored, with SAM 2 masks prompted once with the true mask. $^\dagger$Base read through the top panel, the tube's true shape given. Certified parts leave nine end-panel corners, and on the small carton 13 corners and four hinge readings. Other depth sources and 95\,\% intervals: supplementary. Bold: best, underlined: second, per carton, certified parts set apart.}
  \label{tab:real}
\end{table}

\begin{figure*}[t]
  \centering
  \includegraphics[width=0.9\textwidth,trim=0 22 0 13,clip]{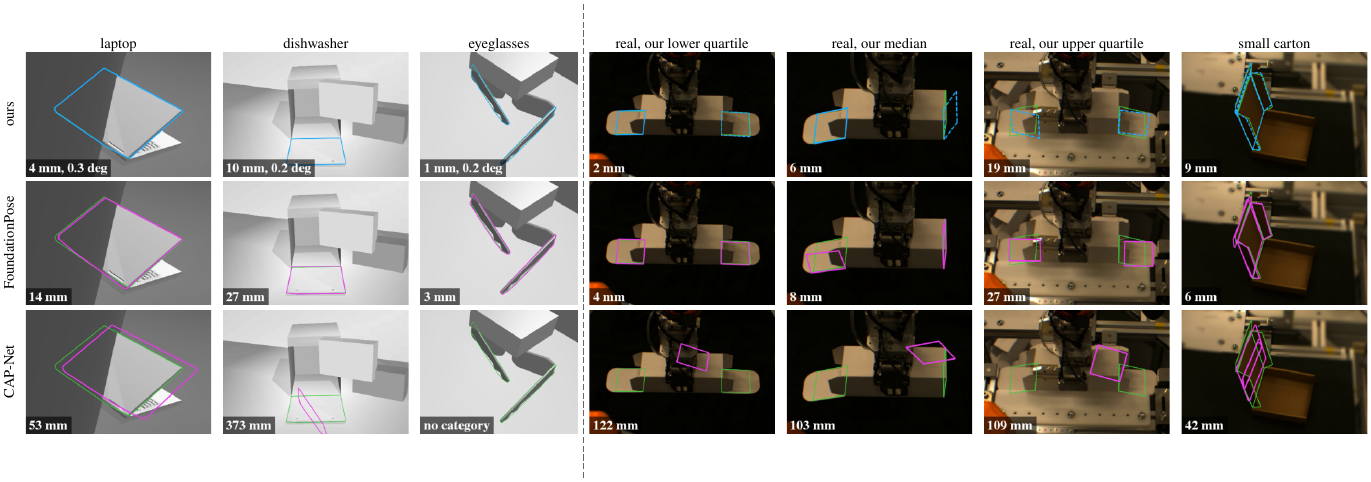}
  \caption{\textbf{Qualitative comparison}, simulated objects partly hidden by other objects (left) and real test frames of the two cartons (right). Truth in green, ours in blue (dashed where the part certificate rejects the part), the baselines, given depth and true masks, in magenta. On the carton we draw only F1 and F3, on the small carton its lid and flaps.
  Under each tile: centre and hinge error. We choose the frames by rule (see supp.).}
  \label{fig:qualitative}
\end{figure*}

From the two colour images alone, our method places the end panels 3.5\,mm off and reads the fold $3.9^\circ$ off (\cref{tab:real},
\cref{fig:qualitative} right). FoundationPose, given depth from the same pair and true masks, reaches 6.3\,mm on learned depth, 5.9\,mm on semi-global depth (see supp.), and
$7.5^\circ$ on the fold only with the tube's true shape given. DART, given the same
depth and mask and a start near the truth, reaches 6.7\,mm and $4.7^\circ$. Under
resampling of the frames, both gaps on the end panels hold, by 1.6 to 12\,mm to FoundationPose on learned depth and narrowly, by 0.1 to 5.0\,mm, to DART, and the fold ties with DART's ($-2.2$ to $2.8^\circ$). DART's tail is shorter than ours over
all frames, though not than that of our certified parts. Given instead one start per recording and masks from SAM 2~\cite{ravi2024sam2}, DART loses the carton on 25 of 44 frames once the masks
spread to a second carton or the jig.
Part masks from SAM 2, prompted the same way, take FoundationPose on learned depth from 6.3 to 27\,mm. Where their IoU with the true masks reaches 0.3, it holds its accuracy (see supp.).
With true masks, FoundationPose wins one median, on the tabs (2.7 against 5.4\,mm), helped by the true mask of a small flat part, with a tail three
times longer. Started from our answer and given true masks, DART shortens the tail of our corners, though not of the fold (see supp.).

Tracking shortens our tail on the end panels, from 99 to 59\,mm, and on the fold, from $167^\circ$ to $82^\circ$. On
five frames the whole fit is lost, and the frame certificate rejects all five. A single wrong panel escapes it, but not the
part certificate (\cref{sec:certificates}), which keeps 16 of the 90 end-panel readings. Requiring fewer certified corners keeps up to 31, three of them more than 10\,mm off (see supp.). Over the full recordings, 13\,\% of end-panel readings are certified on their own frame, but an end panel has a certified angle less than 2\,s old on 48\,\% of frames (see supp.).

A second, smaller carton, a $100\times100\times22$\,mm tray with a lid, two lid flaps and a front flap (\cref{fig:teaser}), tests the same recipe on another real object. Training adds 21 hand-labelled frames and the certified labels of the same two recordings' unlabelled video, and a third recording of the same day, never trained on, gives 12 test frames. From the images alone, we place the moving parts closer than every baseline and read their hinges to $4.0^\circ$ (\cref{tab:real}). DART, given true masks and a start near the truth, reads the hinges better, mostly those of the small lid flaps, and the part certificate keeps only four hinge readings.

\subsection{Ablations}
\label{sec:ablations}
\label{sec:exp_data}
\label{sec:exp_labels}

We switch off one component at a time on the real test session (\cref{tab:ablation}). We compare stages and training data on
single frames, so that tracking does not hide a difference, and the certificates on our full tracked estimates.

\begin{table}[t]
  \centering
  \footnotesize
  \setlength{\tabcolsep}{2pt}
  \begin{tabular}{lccccc}
    \toprule
    variant & frames & cert. & end panels (mm) & tabs (mm) & fold ($^\circ$) \\
    \midrule
    \multicolumn{6}{l}{\emph{(a) stages, single frame}} \\
    A                & 45 & 21 & 15 / 116 & 30 / 122 & 14 / 161 \\
    A+B              & 45 & 27 & 8.8 / 81 & 12 / 81 & 7.5 / \textbf{124} \\
    A+B+C            & 45 & \textbf{28} & 7.1 / 108 & 10 / 94 & 6.1 / 167 \\
    A+B+edges        & 45 & 27 & 4.5 / \textbf{74} & 7.9 / \textbf{72} & 6.0 / 126 \\
    A+B+C+edges      & 45 & \textbf{28} & \textbf{4.1} / 99 & \textbf{6.4} / 90 & \textbf{4.8} / 167 \\
    \midrule
    \multicolumn{6}{l}{\emph{(b) certificates, tracked}} \\
    none             & 45 & -- & 3.5 / 59 & 5.4 / 46 & 3.9 / 82 \\
    frame            & 33 & -- & \textbf{3.1} / 18 & \textbf{5.2} / 19 & \textbf{3.1} / 19 \\
    frame + part     & 33 & -- & 3.7 / \textbf{6.6} & \textbf{5.2} / \textbf{7.1} & 3.7 / \textbf{7.1} \\
    \midrule
    \multicolumn{6}{l}{\emph{(c) detector (A+B) trained on}} \\
    hand labels only & 37 & 0  & 296 / 542 & 294 / 541 & -- \\
    renders only     & 39 & 10 & 108 / 294 & 113 / 316 & 53 / 171 \\
    renders + labels & 45 & 13 & 107 / 312 & 111 / 375 & 26 / 166 \\
    \;+ self-labels  & 45 & \textbf{27} & \textbf{8.8} / \textbf{81} & \textbf{12} / \textbf{81} & \textbf{7.5} / \textbf{124} \\
    \bottomrule
  \end{tabular}
  \caption{\textbf{Ablations} on the real test session, in the columns of \cref{tab:real}. (a) The pipeline's stages. (b) Tracked estimates over the frames they answer, the frames the certificate accepts and the parts it accepts. (c) Stages A and B by training data. Bold: best per block and column.}
  \label{tab:ablation}
\end{table}

\paragraph{Stages.}
Each stage lowers the median error (\cref{tab:ablation}a). Stage B almost halves Stage A's end-panel error and fold, and edge
alignment halves the end-panel error again. Stage C removes the last frames turned end for end. It also lengthens the tails (108 against 81\,mm, $167^\circ$ against $124^\circ$). The corner tail grows only on frames the certificate rejects, where the fit lands in a different minimum, and the fold tail by two end panels read on the wrong side of their hinge, which the part certificate rejects.
In simulation, the fit given exact corners reaches 0.14\,mm, so the error comes from the detector (see supp.).

Without the articulated fit, we cannot measure the fold. Fitting each part rigidly to its own corners answers 24 of 47 frames, 4.9 against our 3.3\,mm on the same corners, and the base panel never has the three clean corners it needs.

\paragraph{Certificates.}
The certificates change no estimate, so they shorten the tail and leave the median about where it was (\cref{tab:ablation}b). The frame certificate removes frames whose fit is lost, which cuts the end panels' 90th
percentile from 59 to 18\,mm, and the part certificate then removes wrong parts on frames that pass, which brings it to 6.6\,mm.

\paragraph{Training data.}
We train Stages A and B from the same initialisation on four sets of data (\cref{tab:ablation}c). Neither source works alone. On
the 30 hand-labelled frames alone, the detector certifies no test frame, and its fit lies 30\,cm from the carton. On renders alone, it
places corners to 2.8 pixels on held-out renders but certifies only 10 test frames, 11\,cm off. Both together answer more frames but
stay 11\,cm off. Self-labels on unlabelled video are the largest single step, bringing the end panels to 8.8\,mm and the
certified frames to 27. We trained this model before we labelled the test session. Two later repeats lose 6 and 11 frames against its 7, none of them certified, and are as precise on the frames they certify (see supp.). On the training videos, one self-label in four lies over 10 pixels off on frames the certificate rejects, against one in forty among those it keeps (see supp.).

\section{Limitations}
\label{sec:limitations}

Our method is instance-level. It needs the object's CAD model, whereas the category-level baselines solve a different problem. It
locates parts through distinct features, such as corners, so it degrades when too many of them are hidden, a problem addressed by optimal task-specific camera placement. Smooth shapes such as spheres lie
outside its scope. For an object with exact symmetry, like the box, we estimate the state up to that symmetry. The baseline of one stereo pair limits its precision, and more cameras help (see supplementary material). We tested only using a widely used consumer-grade depth camera. The fit keeps a hidden part at its last certified angle, so it misses any motion of that part until the part is seen again. When a real hand hides corners of the small carton, the part certificate stays reliable, rejecting every estimate more than 10\,mm off, but it certifies parts on only half as many frames (see supplementary material). We measure the state that reveals a failure but do not evaluate failure detection itself. Finally, the certificates
guarantee consistency with the images and the model, not correctness, and a bias shared by both views passes the epipolar test.

\section{Conclusion}
\label{sec:conclusion}

We recover the articulation state of a known object made of thin, textureless parts from passive stereo,
without a depth sensor. The CAD model is fitted to detected corners and edges, and certificates decide, part by part, which estimates to trust. Certified estimates
on unlabelled video adapt the detector to real images. On a real folding carton with eleven hinges, we place the end panels and
read the fold more precisely than learned estimators given depth and true masks. Given only another object's CAD, our method
locates the certified moving parts of five of the six other objects several times more precisely than the strongest learned baseline on the
same readings. These results show that an exact model and learned corners rival the precision of depth-based fitting from the same two cameras, without its true masks or start, that accurate depth can then refine them, and that certification extends to each part of an articulated object.
Detecting failed operations and estimating material properties from the fitted state are natural next steps.

\section*{LLM Usage Statement}
In this work we used Claude as a coding assistant and writing aid. It helped search for related work, and we read and checked every cited paper ourselves. It wrote code for our experiments, helped set up the baselines' released code, and ran scripts for training, evaluation and figures under our direction. It wrote the scripts that render our synthetic datasets from the carton CAD model and jig that we made and from PartNet-Mobility models, and the tools we used to label real images, while every hand label is ours. It helped refine author-produced results, such as checking scores and analysing failure cases. It generated the figures from our data and to our design, and we reviewed and edited each of them. In the paper, it rewrote paragraphs following our results and instructions, and helped format tables and equations in LaTeX. We chose the method, the experiments and the claims, verified every result and passage it helped produce, and take full responsibility for the final content.

{
    \small
    \bibliographystyle{ieeenat_fullname}
    \bibliography{refs}
}

\end{document}